\PassOptionsToPackage{table}{xcolor}
\documentclass[11pt]{article}

\usepackage[margin=1in]{geometry}
\usepackage{times}
\usepackage[numbers,sort&compress]{natbib}

\usepackage{amsmath}
\usepackage{amssymb}
\usepackage{booktabs}
\usepackage{float}
\usepackage{graphicx}
\usepackage{microtype}
\usepackage{multirow}
\usepackage{placeins}
\usepackage{hyperref}
\usepackage{url}

\usepackage{natbib}
\usepackage{amsthm}

\theoremstyle{remark}

\usepackage{wrapfig}

\newcommand{\best}[1]{\textbf{#1}}
\newcommand{\second}[1]{\underline{#1}}

\usepackage{tabularx}
\usepackage{array}
\usepackage[table]{xcolor}

\definecolor{promptheader}{RGB}{45,45,45}
\definecolor{promptbg}{RGB}{248,248,248}

\usepackage{booktabs}

\usepackage{algorithm}
\usepackage{algpseudocode}

\usepackage{enumitem,listings,needspace}
\usepackage{tcolorbox}
\tcbuselibrary{skins,breakable}

\definecolor{taspoPromptInk}{HTML}{244C4A}
\definecolor{taspoPromptBorder}{HTML}{B9C9C7}
\definecolor{taspoPromptHeader}{HTML}{EDF3F2}
\definecolor{taspoPromptPaper}{HTML}{FAFBFB}
\definecolor{TASPOgreen}{HTML}{E8F3F0}
\definecolor{modelgray}{HTML}{F1F3F5}
\newtcolorbox{taspoPromptBox}[1]{
  enhanced,
  breakable,
  colback=taspoPromptPaper,
  colframe=taspoPromptBorder,
  colbacktitle=taspoPromptHeader,
  coltitle=taspoPromptInk,
  fonttitle=\small\sffamily\bfseries,
  fontupper=\small\normalfont,
  title={#1},
  title after break={#1\enspace(continued)},
  boxrule=0.45pt,
  arc=1.2mm,
  outer arc=1.2mm,
  boxsep=0pt,
  left=8pt,
  right=8pt,
  top=6pt,
  bottom=6pt,
  toptitle=4pt,
  bottomtitle=4pt,
  before skip=9pt,
  after skip=9pt,
  before upper={\setlength{\parindent}{0pt}}
}

\lstdefinestyle{taspoSchema}{
  basicstyle=\ttfamily\fontsize{8.3}{10.2}\selectfont,
  columns=fullflexible,
  keepspaces=true,
  showstringspaces=false,
  breaklines=true,
  breakatwhitespace=false,
  breakindent=1.5em,
  tabsize=2,
  aboveskip=4pt,
  belowskip=0pt,
  numbers=none,
  frame=none
}

\usepackage{amsmath,amsfonts,bm}

\def\eqref#1{equation~\ref{#1}}
\def\1{\bm{1}}

\DeclareMathAlphabet{\mathsfit}{\encodingdefault}{\sfdefault}{m}{sl}
\SetMathAlphabet{\mathsfit}{bold}{\encodingdefault}{\sfdefault}{bx}{n}

\title{Reconciling Process Supervision with Outcome-Based Credit in Agentic Policy Optimization
}

\author{
Jingxiao Yang$^{1}$\thanks{Equal contribution.},
Wangjie Gan$^{1}$\footnotemark[1],
Yingxuan Zhuang$^{1}$\footnotemark[1],
Wenqi Zhang$^{1}$,
Jintao Chen$^{1}$,
Xuhong Zhang$^{1}$
\\
\normalsize $^{1}$ Zhejiang University
}

\date{}

\begin{document}

\maketitle
% \iclrfinalcopy

\begin{abstract}
For language-model agents, outcome-based methods such as GRPO use verified trajectory feedback but assign the same advantage to all intermediate decisions, obscuring their different contributions to success or failure. On-policy self-distillation supplies denser process supervision by re-evaluating sampled behavior with training-time privileged information (PI). Yet finer supervision does not automatically provide outcome-consistent credit. PI-induced likelihood changes reflect context-dependent policy preferences, but do not specify how outcome-derived advantages should be distributed across individual interaction decisions. Directly optimizing these signals can introduce objectives that compete with outcome-based reinforcement. Their usefulness also depends on whether the guidance applies to the target interaction and whether token-level changes meaningfully evaluate complete executable actions.
We introduce TASPO, a framework for reconciling process supervision with outcome-based credit in agentic policy optimization. TASPO aligns guidance from verified successful sibling trajectories with the target interaction and aggregates PI-induced likelihood shifts over executable actions. The resulting relative support reweights the original advantage through positive, bounded, mean-one weights, preserving its sign and trajectory mean without additional environment interaction.
Across ALFWorld, Search-QA, and WebShop with three backbone models, TASPO achieves an average gain of 10.6\% over GRPO under same interaction budgets and policy-update counts. Analyses favor aligned supervision and indicate greater training stability with action-level allocation.

\end{abstract}

\section{Introduction}
Reinforcement learning trains language-model agents to solve multi-turn tasks through interaction with external environments \citep{jin2025searchr1,feng2025retool,wang2025ragen}. Outcome-based methods use verified trajectory feedback, but with terminal rewards, GRPO assigns the resulting group-relative advantage uniformly to intermediate decisions \citep{shao2024deepseekmath}. Consequently, useful and counterproductive actions within a trajectory receive identical advantage values. Process supervision provides information about intermediate behavior, creating an opportunity to refine this coarse assignment. The challenge is to integrate such information with outcome-based credit for more effective agentic policy optimization.

Existing approaches refine intermediate credit using environment-state structure or learned process evaluators and value estimates \citep{lightman2024verify, feng2025gigpo,choudhury2025process}. On-policy (self)-distillation (OPSD)\citep{agarwal2024opd,zhang2026stepopsd,zhao2026selfdistilledreasoner,lu2026sdar,yang2026opid} offers another route: a teacher branch of the model conditions on privileged information (PI) available only during training to supervise responses sampled without PI. This mechanism turns additional context into dense token-level guidance without requiring a separately trained process reward model. OPSD thus provides a practical mechanism for incorporating process information into outcome-based agent training. However, integrating such supervision with outcome-based credit presents two challenges.

First, process supervision and outcome-based reinforcement can impose different optimization objectives. In OPSD, PI-induced likelihood shifts describe how additional context changes policy preference, they do not specify how each decision should share the trajectory's outcome-derived advantage. As illustrated in Figure~\ref{fig:gradient_geometry}, adding a distillation loss introduces an independent preference-matching gradient that need not agree with the outcome-derived policy gradient, even when the process signal is coherent. Weighting or gating this auxiliary objective regulates its influence \citep{lu2026sdar,zhang2026stepopsd,ding2026sgcd}, but does not by itself establish an outcome-consistent allocation of local credit. This motivates integrating process supervision through constrained redistribution of outcome-derived advantages, retaining the outcome as the basis for credit while allowing local information to refine its allocation.

\begin{figure*}[t]
    \centering
    \includegraphics[width=\textwidth]{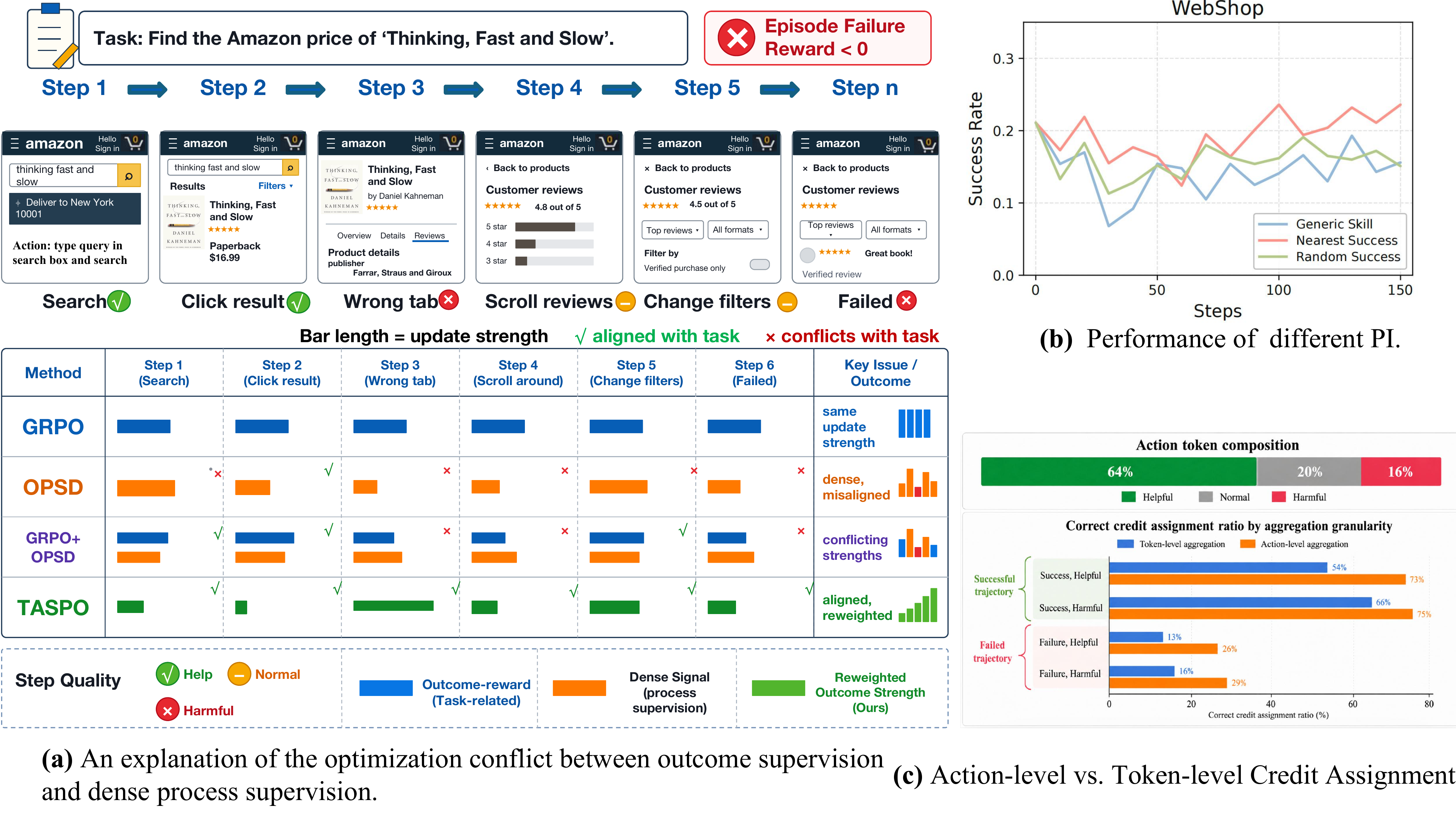}
    \vspace{-0.6em}
    \caption{
    \textbf{An explanation of why privileged process supervision should refine, rather than replace, verified outcome credit.}
    \textbf{(a)} Uniform outcome credit and dense process signals can conflict when directly combined.
    \textbf{(b)} Self-distillation performance depends strongly on the source and relevance of privileged information.
    \textbf{(c)} Action-level aggregation yields more accurate credit assignment than token-level aggregation across trajectory outcomes and action qualities.
    }
    \label{fig:motivation}
    \vspace{-0.8em}
\end{figure*}

Second, process supervision may be misaligned with the interaction decisions to which credit is assigned. In OPSD, task-relevant PI may still be inapplicable to the target trajectory's actual state or execution path\citep{xu2026tip,fu2026revisitingopd}. Moreover, token-wise likelihood shifts describe individual token preferences, whereas the environment executes complete actions. Applying these shifts directly can assign inconsistent credit to tokens within one action. Figure~\ref{fig:motivation}(b,c) highlights the sensitivity to PI selection and supervision granularity. Effective integration therefore calls for supervision applicable to the target interaction and organized around executable actions as coherent decision units.

To address these challenges, we introduce \textbf{TASPO}, a framework for reconciling process supervision with outcome-based credit in agentic policy optimization. TASPO uses process information to refine how outcome-derived advantages are distributed across decisions. It obtains these signals using trajectory-aligned PI: a training-time analyzer extracts conditional guidance from verified successful sibling trajectories within the same rollout group and checks its applicability against each target's recorded action--observation history. Retained guidance forms a shared PI context; if none applies, TASPO falls back to GRPO. The frozen rollout policy scores sampled responses with and without this context, and likelihood shifts are aggregated over executable-action tokens. Relative action support is converted into positive, bounded, mean-one weights on the original advantage. Stronger support increases positive credit or attenuates negative credit. Each weight applies to all policy-generated tokens in its turn, preserving the advantage sign and trajectory mean without additional environment interaction.

\begin{wrapfigure}{r}{0.52\columnwidth}
    \vspace{-3.0em}
    \centering
    \includegraphics[width=\linewidth]{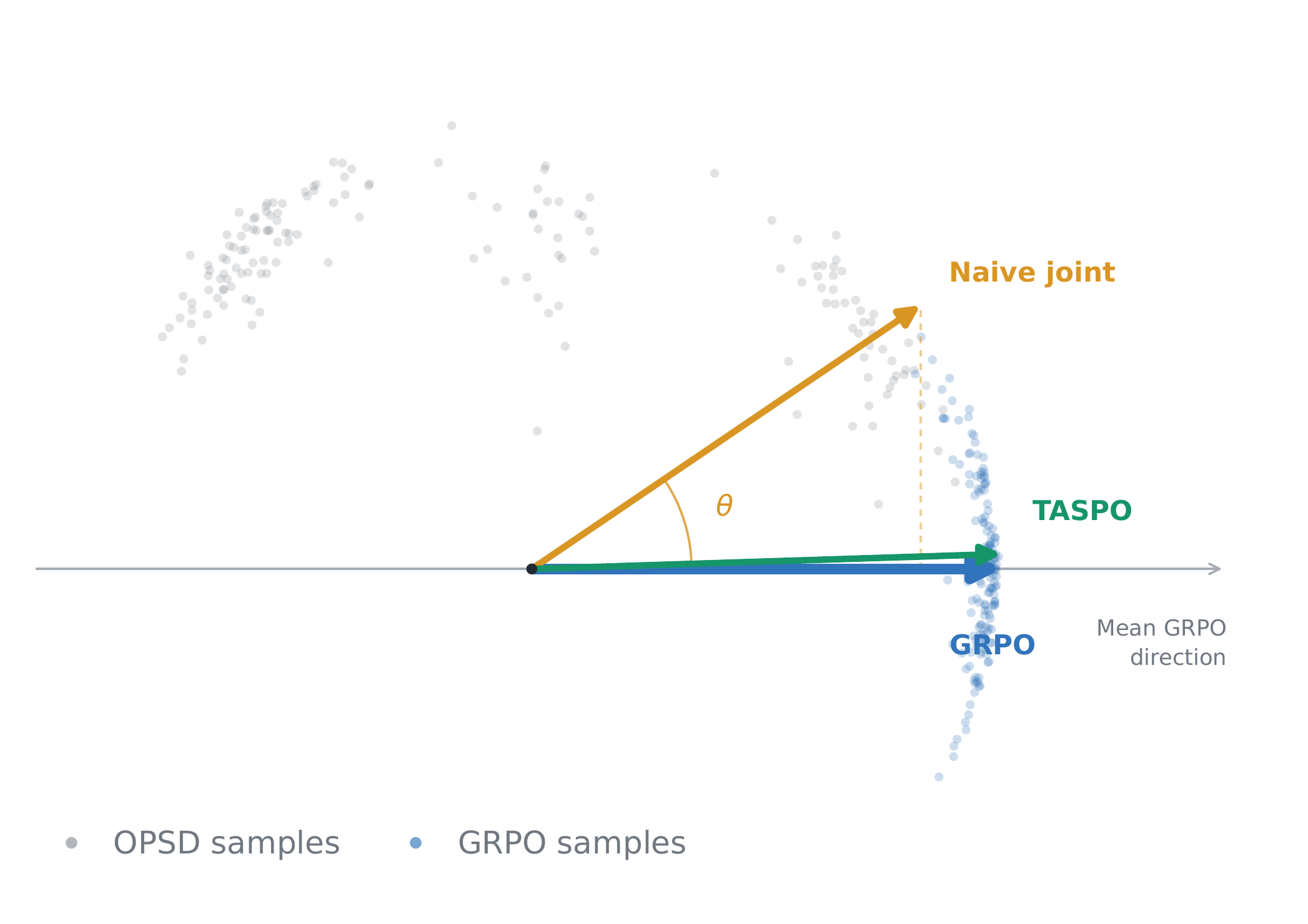}
    \caption{
    \textbf{Why PI should reweight credit rather than steer optimization.} OPSD produces coherent but off-outcome gradients, while TASPO preserves the outcome direction through credit reweighting.
    }
    \label{fig:gradient_geometry}
    \vspace{-1.0em}
\end{wrapfigure}

We evaluate TASPO on ALFWorld, Search-QA\citep{shridhar2020alfworld,yao2022webshop,jin2025searchr1}, and WebShop with three backbone models. Across nine benchmark--backbone settings, TASPO improves over GRPO by an average of 10.6\% under matched interaction budgets and policy-update counts. Analyses of supervision construction favor trajectory-aligned guidance over generic skills and direct trajectory reuse. Action-level allocation improves performance and reduces variation across training seeds, with smoother reward and policy-KL dynamics on ALFWorld. Comparisons with standalone OPSD and GRPO+OPSD on ALFWorld further support the effectiveness of constrained credit redistribution.
Our contributions are:

1. We formulate the integration of process supervision and outcome feedback as a credit-assignment problem, focusing on optimization consistency and decision alignment.

2. We propose TASPO, which integrates trajectory-aligned, action-level process supervision into outcome-based optimization through constrained credit redistribution.

3. We demonstrate consistent improvements over GRPO across three benchmarks and three backbones, supported by analyses of supervision construction, allocation granularity, and optimization dynamics.

% \begin{figure*}[t]
%   \centering
%   \begin{minipage}[t]{0.32\textwidth}
%     \centering
%     \includegraphics[width=\linewidth]{figures/p1_motivation.pdf}
%     \vspace{-0.5ex}

%     \small\textbf{(a)}
%   \end{minipage}\hfill
%   \begin{minipage}[t]{0.32\textwidth}
%     \centering
%     \includegraphics[width=\linewidth]{figures/p1_motivation_new.pdf}
%     \vspace{-0.5ex}

%     \small\textbf{(b)}
%   \end{minipage}\hfill
%   \begin{minipage}[t]{0.32\textwidth}
%     \centering
%     \includegraphics[width=\linewidth]{figures/p3_motivation.png}
%     \vspace{-0.5ex}

%     \small\textbf{(c)} 
%   \end{minipage}
%   \caption{Illustration of our motivating observations. \textbf{(a)} Outcome-level verifiable rewards assign the same advantage to actions of different quality within a trajectory. 
%   \textbf{(b)} The effectiveness of on-policy self-distillation varies with the source of privileged information.
%   \textbf{(c)} OPSD supervision contains useful signals for distinguishing action quality, while action-level aggregation produces credit adjustments more consistent with whether an action is helpful or harmful.
%   }
%   \label{fig:motivation}
% \end{figure*}

\section{Related Work}
\label{sec:related-work}

\subsection{Agentic Reinforcement Learning and Credit Assignment}

Language models are increasingly trained as interactive agents that reason,
search, invoke tools, and act over long horizons
\citep{yao2023react,schick2023toolformer,liu2024agentbench}.
Outcome-based reinforcement learning provides a direct way to optimize such
agents using environment or verifier feedback
\citep{shao2024deepseekmath,guo2025deepseekr1,jin2025searchr1,
feng2025retool,wang2025ragen,dong2025arpo,feng2025gigpo}.
However, sparse trajectory outcomes provide limited information about which
intermediate decisions caused success or failure, reflecting the broader
delayed-credit problem studied in reinforcement and sequential learning
\citep{andrychowicz2017her,arjonamedina2019rudder,lightman2024verify}.
Recent agentic methods refine policy optimization using additional structure
within sampled trajectories \citep{dong2025arpo,feng2025gigpo,
he2026actionbottleneck}. TASPO instead uses training-time privileged
information to differentiate decisions within a trajectory while retaining
the verified outcome as the source of the overall update direction.

\subsection{On-Policy Self-Distillation with Privileged Information}

Knowledge distillation transfers behavioral information from a teacher through
token- or sequence-level supervision
\citep{hinton2015distilling,kim2016sequence}. On-policy distillation further
supervises student-generated samples, reducing distribution mismatch between
the data being optimized and the current policy
\citep{ross2011dagger,agarwal2024opd}. Recent self-distillation methods
construct privileged teachers from additional contexts, reasoning traces,
skills, or hindsight available only during training
\citep{zhao2026selfdistilledreasoner,ye2026opcd,yang2026rlsd, yang2026sare, zhuang2026, 
lu2026sdar,wang2026skillsd,yang2026opid,wu2026seed}.
Several works further adapt such supervision for finer-grained agent credit,
including step-aware, sibling-guided, and counterfactual formulations
\citep{zhang2026stepopsd,ding2026sgcd,meng2026craft}.
A complementary line of work studies when dense on-policy supervision is
reliable, considering teacher--student compatibility, token importance,
trajectory position, curriculum, and trust-region constraints
\citep{fu2026revisitingopd,li2026rethinkingopd,xu2026tip,
xing2026tropd,li2026guidedopd,xie2026positionbias}.
TASPO focuses on two distinct questions: whether trajectory-derived guidance
is applicable to the target decision, and how such guidance should refine
credit without overriding verified outcomes. It therefore aligns
source-grounded PI to each target pre-action context and uses the resulting
signal only to redistribute outcome-derived credit across executable actions.

\section{Method}
\label{sec:method}
\begin{figure*}[t]
  \centering
  \includegraphics[width=\textwidth]{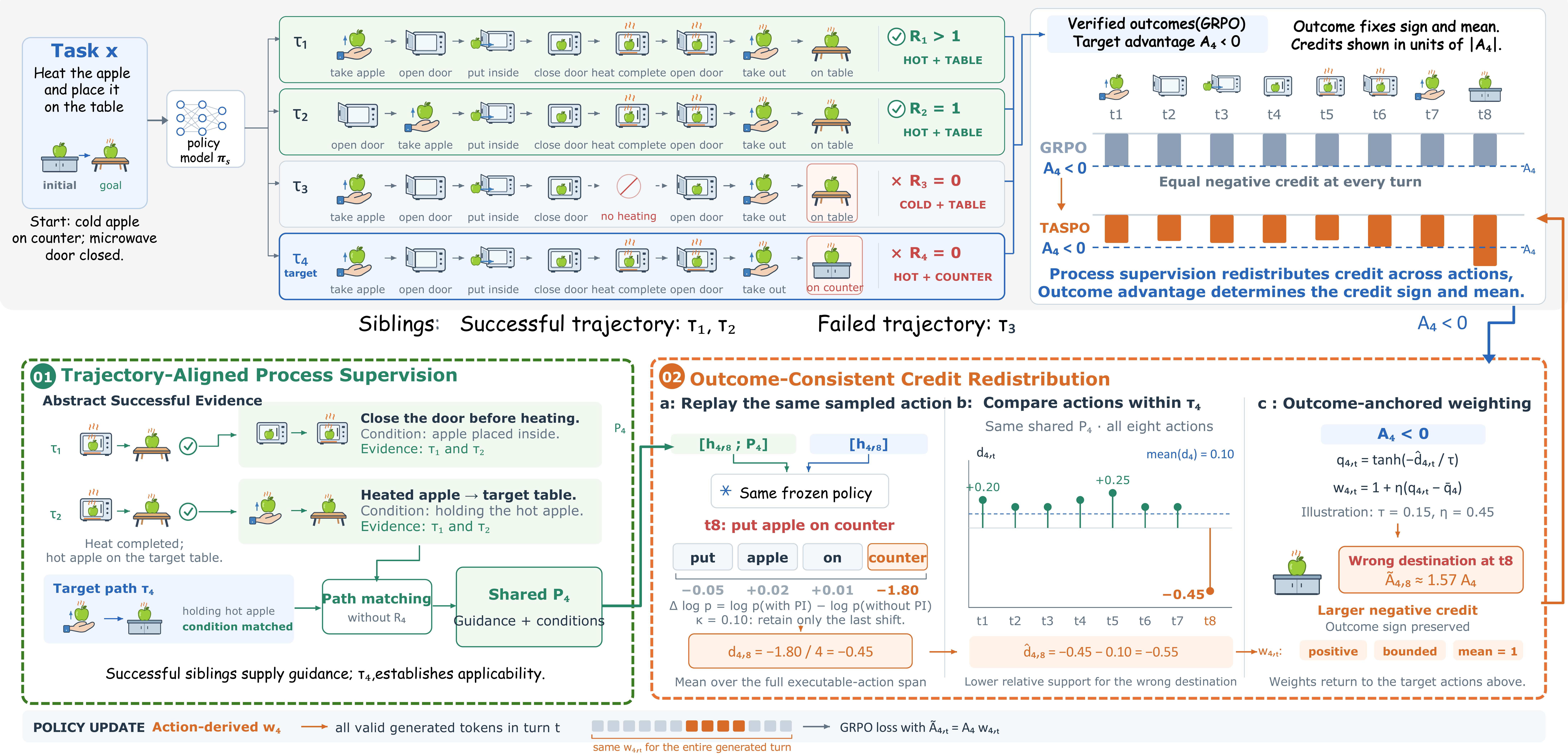}
  \caption{\textbf{Overview of TASPO.} TASPO reconciles process supervision with outcome-based credit using shared, trajectory-aligned PI from verified successful siblings. Action-level likelihood shifts yield positive, bounded, mean-one advantage weights, refining local credit while preserving the original advantage's sign and trajectory mean.
  }
  \label{fig:method-overview}
\end{figure*}

TASPO uses process supervision to distinguish actions that otherwise share the same outcome advantage.
To make this distinction useful for policy optimization, we first specify how process information may modify the advantage, then construct guidance applicable to the target interaction and compare its support across executable actions.
The resulting scores determine the advantage redistribution shown in Figure~\ref{fig:method-overview}.

\subsection{Outcome-Constrained Credit Assignment}
\label{sec:problem-setup}

For a task $x$, the rollout policy $\pi_{\theta_{\mathrm{old}}}$ samples $K$ trajectories $\mathcal G_x=\{\tau_i\}_{i=1}^{K}$.
At turn $t$, it generates response $y_{it}$ from history $h_{it}$, and the environment executes action $a_{it}$.
The verified terminal reward $R_i$ yields the GRPO advantage
$A_i=(R_i-\mu_{\mathcal G_x})/(\sigma_{\mathcal G_x}+\delta)$,
where $\mu_{\mathcal G_x}$ and $\sigma_{\mathcal G_x}$ are the group reward mean and standard deviation, and $\delta>0$ stabilizes normalization.
GRPO uses the same $A_i$ across all $T_i$ turns.
We let process supervision differentiate these turns by reweighting $A_i$ within the trajectory:
\begin{equation}
\widetilde A_{it}=A_iw_{it},
\qquad
1-\epsilon_w\leq w_{it}\leq1+\epsilon_w,
\qquad
\frac{1}{T_i}\sum_{t=1}^{T_i}w_{it}=1,
\label{eq:allocation-goal}
\end{equation}
with $0\leq\epsilon_w<1$.
We retain positive weights so that each turn inherits the original advantage sign, and require mean-one weights to preserve the average advantage.
Guidance can still be applied to unsuitable states, and likelihood increases need not indicate task progress.
We therefore bound each turn's adjustment by $\epsilon_w|A_i|$.
Process supervision thus enters the GRPO advantage without a separate PI-matching objective.
These guarantees concern scalar advantages; the parameter-gradient direction can still change.

\subsection{Trajectory-Aligned Process Supervision}
\label{sec:alignment}

We first need guidance that applies to the target's actual interaction.
A training-time analyzer extracts conditional guidance $\mathcal E_x$ from verified successful trajectories $\mathcal S_x\subseteq\mathcal G_x$, such as task requirements, action preconditions, and ordering constraints.
Each item cites supporting source actions and observations.
These conditions may differ across rollouts, so source evidence alone is insufficient to establish applicability.

For target $\tau_i$, we exclude items supported only by that target to obtain $\mathcal E_{x,-i}$.
The analyzer then checks the remaining guidance against the task and the target's complete recorded action--observation history $\mathcal H_i$.
Source interactions support the guidance itself; target interactions provide the basis for assessing its relevance.
Retained or reformulated statements cite both.
Alignment occurs after rollout without receiving the target's explicit terminal reward or success label.
The resulting statements form a compact privileged context:
\begin{equation}
P_i=
\operatorname{Compose}\!\left(
\operatorname{Align}(\mathcal E_{x,-i},x,\mathcal H_i)
\right).
\label{eq:trajectory-pi}
\end{equation}
We reuse this $P_i$ across turns to keep the guidance fixed during action comparison, while retaining each turn's original history and response prefix.
If no statements pass the evidence checks, $P_i$ is empty.
The appendix details the prompts, evidence checks, and context budget.

\subsection{Trajectory-Relative Action Scoring}
\label{sec:relative-scoring}

With $P_i$ fixed, we use the privileged rescoring mechanism of OPSD to measure how the guidance changes the likelihood of sampled behavior.
We focus on the tokens specifying the action submitted to the environment, so that the scoring unit matches the decision whose credit we adjust.
Let $\mathcal M_{it}$ denote these token positions in $y_{it}$.
The frozen rollout policy computes
\begin{equation}
\Delta_{itk}
=
\operatorname{sg}\!\left[
\log\pi_{\theta_{\mathrm{old}}}
(y_{itk}\mid P_i,h_{it},y_{it,<k})
-
\log\pi_{\theta_{\mathrm{old}}}
(y_{itk}\mid h_{it},y_{it,<k})
\right],
\label{eq:token-shift}
\end{equation}
where $\operatorname{sg}$ stops gradients.
A positive $\Delta_{itk}$ means that adding PI makes the sampled token more likely.
Let $\mathcal I_i$ contain turns with available scoring inputs, a complete nonempty action span, and finite log probabilities and shifts for every action token.

A practical issue is that a single extreme shift can dominate the action mean and distort comparisons with other actions.
We therefore clip token shifts before averaging and center the resulting action scores:
\begin{equation}
\begin{aligned}
d_{it}
&=
\frac{1}{|\mathcal M_{it}|}
\sum_{k\in\mathcal M_{it}}
\operatorname{clip}(\Delta_{itk},-c_\Delta,c_\Delta),\\
\widehat d_{it}
&=
d_{it}-
\frac{1}{|\mathcal I_i|}
\sum_{u\in\mathcal I_i}d_{iu},
\qquad t\in\mathcal I_i.
\end{aligned}
\label{eq:relative-action-shift}
\end{equation}
Here, $c_\Delta>0$ limits individual token shifts, and averaging avoids directly scaling the score with action length.
Centering provides a reference within the trajectory: $\widehat d_{it}>0$ means the action's score exceeds the mean over scorable turns.
We use this relative support to allocate the outcome advantage.

\subsection{Advantage Redistribution and Policy Optimization}
\label{sec:credit-allocation}

We note that on a negative-advantage trajectory, larger weights produce more negative credit.
Simply increasing weights with PI support would therefore penalize better-supported actions more heavily.
To obtain the intended allocation for both advantage signs, we use
\begin{equation}
q_{it}
=
\tanh\!\left(
\frac{\operatorname{sign}(A_i)\widehat d_{it}}{\tau}
\right),
\qquad t\in\mathcal I_i,
\label{eq:outcome-oriented-score}
\end{equation}
for $A_i\neq0$ and $|\mathcal I_i|\geq2$, where $\tau>0$ controls sensitivity.
We then convert these scores into deviations from unit weights:
\begin{equation}
w_{it}=1+\eta(q_{it}-\bar q_i),
\qquad
\bar q_i=\frac{1}{|\mathcal I_i|}\sum_{u\in\mathcal I_i}q_{iu},
\qquad
\eta=\frac{\epsilon_w}{2}.
\label{eq:action-weight}
\end{equation}
The second centering is necessary because $\tanh$ need not preserve the zero mean of $\widehat d_{it}$.
It makes the weights average to one over $\mathcal I_i$, while $q_{it}\in[-1,1]$ gives $|w_{it}-1|\leq\epsilon_w$.
This favors better-supported actions with relatively more positive credit when $A_i>0$ and less negative credit when $A_i<0$.
The appendix provides the derivation.

Unscorable turns retain $w_{it}=1$, allowing the remaining turns to participate.
If shared PI cannot be constructed, $A_i=0$, or fewer than two turns are scorable, all weights remain one; fewer than two scores provide no relative comparison.
Since excluded turns retain unit weights, redistribution within $\mathcal I_i$ also preserves the mean over the complete trajectory, satisfying Eq.~\ref{eq:allocation-goal}.

Finally, we apply each action-derived advantage to the entire response that produced the action.
Let $\mathcal R_{it}$ contain policy-generated token positions, including reasoning and action tokens but excluding environment observations.
The policy objective is
\begin{equation}
\mathcal L_{\mathrm{policy}}
=
\frac{1}{B}\sum_{i=1}^{B}
\frac{1}{T_i}\sum_{t=1}^{T_i}
\frac{1}{|\mathcal R_{it}|}
\sum_{k\in\mathcal R_{it}}
\ell^{\mathrm{GRPO}}_{itk}(\theta;\widetilde A_{it}),
\label{eq:hierarchical-objective}
\end{equation}
where $B$ is the number of trajectories and $\ell^{\mathrm{GRPO}}$ is the GRPO token policy loss with the supplied advantage.
Averaging within turns and then across turns gives each turn equal base weight before redistribution.
Any KL regularization follows the GRPO baseline.
PI construction and rescoring use completed rollouts during training and require no additional environment interaction.

\section{Experiments}
\label{sec:experiments}

% Our experiments examine both whether TASPO improves long-horizon agent performance and whether the gains arise from its two central design choices: decision-level PI alignment and outcome-anchored credit allocation. We first compare downstream performance under matched environment-interaction budgets, including generalization to held-out environments. We then analyze whether trajectory alignment filters guidance that is incompatible with the current decision context and whether action-level scoring produces credit adjustments that better reflect action quality than token-level supervision. Finally, controlled ablations isolate the contribution of each component.

\begin{table*}[t]
\centering
\caption{
Main results on ALFWorld, Search-QA, and WebShop across different backbone models.
ALFWorld reports success rate (\%) over six task families, Search-QA reports performance over seven QA datasets, and WebShop reports Score and success rate (Succ.).
Avg. denotes the reported aggregate performance on the corresponding benchmark.
\textbf{Bold} and \underline{underlined} values indicate the best and second-best results within each backbone, respectively; ties receive the same formatting.
}
\label{tab:main-results}

\setlength{\tabcolsep}{3.5pt}
\renewcommand{\arraystretch}{1.08}

\resizebox{\textwidth}{!}{
\begin{tabular}{
l
rrrrrrr
rrrrrrrr
rr
}
\toprule
& \multicolumn{7}{c}{\textbf{ALFWorld}}
& \multicolumn{8}{c}{\textbf{Search-QA}}
& \multicolumn{2}{c}{\textbf{WebShop}}
\\
\cmidrule(lr){2-8}
\cmidrule(lr){9-16}
\cmidrule(lr){17-18}

\textbf{Method}
& Pick & Look & Clean & Heat & Cool & Pick2 & \textbf{Avg.}
& NQ & Triv & Pop & Hotp & 2Wk & MuS & Bam & \textbf{Avg.}
& Score & Succ.
\\
\midrule

% =========================================================
% Qwen2.5-3B-Instruct
% =========================================================
\multicolumn{18}{l}{\textit{Qwen2.5-3B-Instruct}}\\[-1pt]

Vanilla
& 44.4 & 11.1 & 6.2 & 15.4 & 28.6 & 12.5 & 21.9
& 24.6 & 48.1 & 31.0 & 26.3 & 25.3 & 7.2 & 59.7 & 31.7
& 6.7 & 0.8
\\

OPSD
& 48.8 & 41.7 & 16.7 & 0.0 & 15.8 & 16.7 & 28.1
& 0.1 & 0.1 & 0.1 & 0.0 & 0.0 & 0.0 & 0.0 & 0.0
& 11.3 & 3.1
\\

GRPO
& 87.9 & 68.8 & 81.8 & 58.3 & 65.4 & 68.4 & 74.2
& 39.3 & 60.6 & 41.1 & 37.4 & 34.6 & 15.4 & 26.4 & 36.4
& 79.8 & 63.3
\\

SDAR
& \second{97.1} & 62.5 & \best{100.0} & 61.9 & 75.0 & \second{84.2} & \second{84.4}
& 44.8 & 58.1 & 44.3 & 38.6 & 36.2 & 15.7 & \best{66.1} & 43.4
& \second{85.0} & 68.0
\\

StepOPSD
& \second{97.1} & 66.7 & 87.0 & \second{79.1} & 78.9 & \best{95.0} & 83.6
& 43.6 & 61.2 & 43.8 & 39.2 & \second{38.1} & 15.8 & \second{64.5} & \second{43.7}
& -- & --
\\

OPID
& 92.7 & \best{100.0} & 88.9 & 70.0 & \best{84.2} & 70.0 & 84.3
& \second{45.9} & \second{61.4} & \best{45.7} & \second{40.7} & \best{38.8} & \second{16.4} & \best{66.1} & \best{45.0}
& \second{85.0} & \second{74.2}
\\

\textbf{TASPO}
& \best{97.9} & \second{77.5} & \second{95.8} & \best{95.8} & \second{79.6} & 70.8 & \best{86.3}
& \best{46.8} & \best{65.5} & \second{44.9} & \best{41.1} & 31.4 & \best{19.1} & 55.2 & 43.4
& \best{88.5} & \best{78.1}
\\

\midrule

% =========================================================
% Qwen2.5-7B-Instruct
% =========================================================
\multicolumn{18}{l}{\textit{Qwen2.5-7B-Instruct}}\\[-1pt]

Vanilla
& 36.1 & 22.2 & 3.1 & 0.0 & 0.0 & 0.0 & 12.5
& 25.2 & 50.8 & 29.5 & 29.0 & 29.0 & 10.4 & 63.7 & 33.9
& 5.9 & 1.6
\\

GRPO
& 91.2 & 70.0 & 90.9 & 78.6 & 64.7 & 66.7 & 78.9
& 46.3 & 63.8 & 47.2 & 44.6 & \second{44.3} & 19.3 & 71.4 & 48.6
& 85.7 & 75.0
\\

OPSD
& 50.0 & 60.0 & 22.7 & 21.4 & 17.6 & 9.5 & 32.8
& 8.8 & 8.6 & 17.5 & 2.5 & 4.2 & 0.5 & 1.2 & 6.2
& 4.5 & 2.3
\\

SDAR
& 94.7 & 75.0 & \best{100.0} & 86.7 & 68.2 & 78.9 & \second{85.9}
& 46.3 & 63.5 & \second{48.2} & 43.8 & \best{48.4} & \second{19.6} & \second{73.0} & 49.0
& \best{89.4} & \best{82.8}
\\

OPID
& \best{100.0} & \second{81.8} & \second{97.1} & \best{100.0} & \second{80.8} & \second{80.0} & \best{90.0}
& \second{48.8} & \second{65.6} & 46.8 & \best{46.1} & 42.7 & \best{21.7} & 72.6 & \second{49.2}
& 85.3 & \second{79.7}
\\

\textbf{TASPO}
& \second{98.2} & \best{84.8} & 94.4 & \second{98.3} & \best{83.3} & \best{80.7} & \best{90.0}
& \best{49.6} & \best{65.9} & \best{49.3} & \second{45.1} & 43.3 & 18.4 & \best{76.7} & \best{49.8}
& \second{87.7} & 76.4
\\

\midrule

% =========================================================
% Qwen3-1.7B-Instruct
% =========================================================
\multicolumn{18}{l}{\textit{Qwen3-1.7B-Instruct}}\\[-1pt]

Vanilla
& 25.0 & 22.2 & 3.1 & 0.0 & 21.4 & 4.2 & 12.5
& 29.4 & 46.9 & 37.0 & 23.5 & 19.6 & 6.4 & 10.5 & 24.8
& 46.5 & 4.7
\\

OPSD
& 26.3 & 33.3 & 9.1 & 0.0 & 4.5 & 5.3 & 14.1 
& 4.2 & 8.3 & 4.6 & 6.6 & 15.3 & 0.7 & 1.2 & 5.8
& 47.4 & 9.3
\\

GRPO
& 59.3 & 46.7 & 12.5 & \best{68.4} & 16.7 & 36.8 & 39.1
& 43.0 & 59.7 & \best{47.4} & 37.9 & 37.4 & 13.6 & 66.1 & 44.3
& 55.8 & 43.0
\\

SDAR
& \best{73.5} & 25.0 & \best{76.9} & 33.3 & 40.0 & 36.8 & 53.9
& 39.7 & 58.9 & 45.3 & 35.9 & 35.5 & 12.6 & 65.3 & 41.9
& 76.8 & 58.6
\\

StepOPSD
& 64.7 & 44.4 & 56.5 & \second{60.9} & 42.1 & \second{55.0} & 56.3
& 40.5 & 58.6 & 45.6 & 34.9 & 29.8 & 10.8 & 64.5 & 41.4
& -- & --
\\

OPID
& 65.9 & \best{72.7} & 66.7 & 40.0 & \second{63.2} & 45.0 & \second{58.9}
& \best{48.8} & \best{65.6} & \second{46.8} & \best{46.1} & \second{42.7} & \best{21.7} & \best{72.6} & \best{49.2}
& \best{85.3} & \best{79.7}
\\

\textbf{TASPO}
& \second{71.2} & \second{64.8} & \second{68.4} & 60.3 & \best{73.3} & \best{60.7} & \best{66.5}
& \second{46.1} & \second{62.6} & 45.4 & \second{40.5} & \best{43.7} & \second{19.7} & \second{68.5} & \second{46.7}
& \second{79.6} & \second{64.8}
\\

\bottomrule
\end{tabular}
}

\end{table*}

\subsection{Experimental Setup}
\label{sec:experimental-setup}

\textbf{Benchmarks and evaluation.}
We evaluate TASPO on three long-horizon agent domains: ALFWorld, WebShop, and Search-QA, following evaluation protocols used in recent agentic RL and
distillation studies \citep{lu2026sdar,zhang2026stepopsd,yang2026opid,wu2026seed}. For ALFWorld, we report success rates on the six task families---Pick, Look, Clean, Heat, Cool, and Pick2---together with their macro-average. Main results use the 140-task seen split, and matched reruns are additionally evaluated on the 134-task unseen split. For WebShop, we evaluate on 128 fixed tasks and report normalized task score and success rate. Search-QA comprises NQ, TriviaQA, PopQA, HotpotQA, 2WikiMultiHopQA, MuSiQue, and Bamboogle. We train on NQ and HotpotQA and report exact-match accuracy on all seven datasets and their macro-average.

\begin{wrapfigure}{r}{0.5\columnwidth}
    \centering
    \vspace{-1.0em}
    \includegraphics[width=0.5\columnwidth]{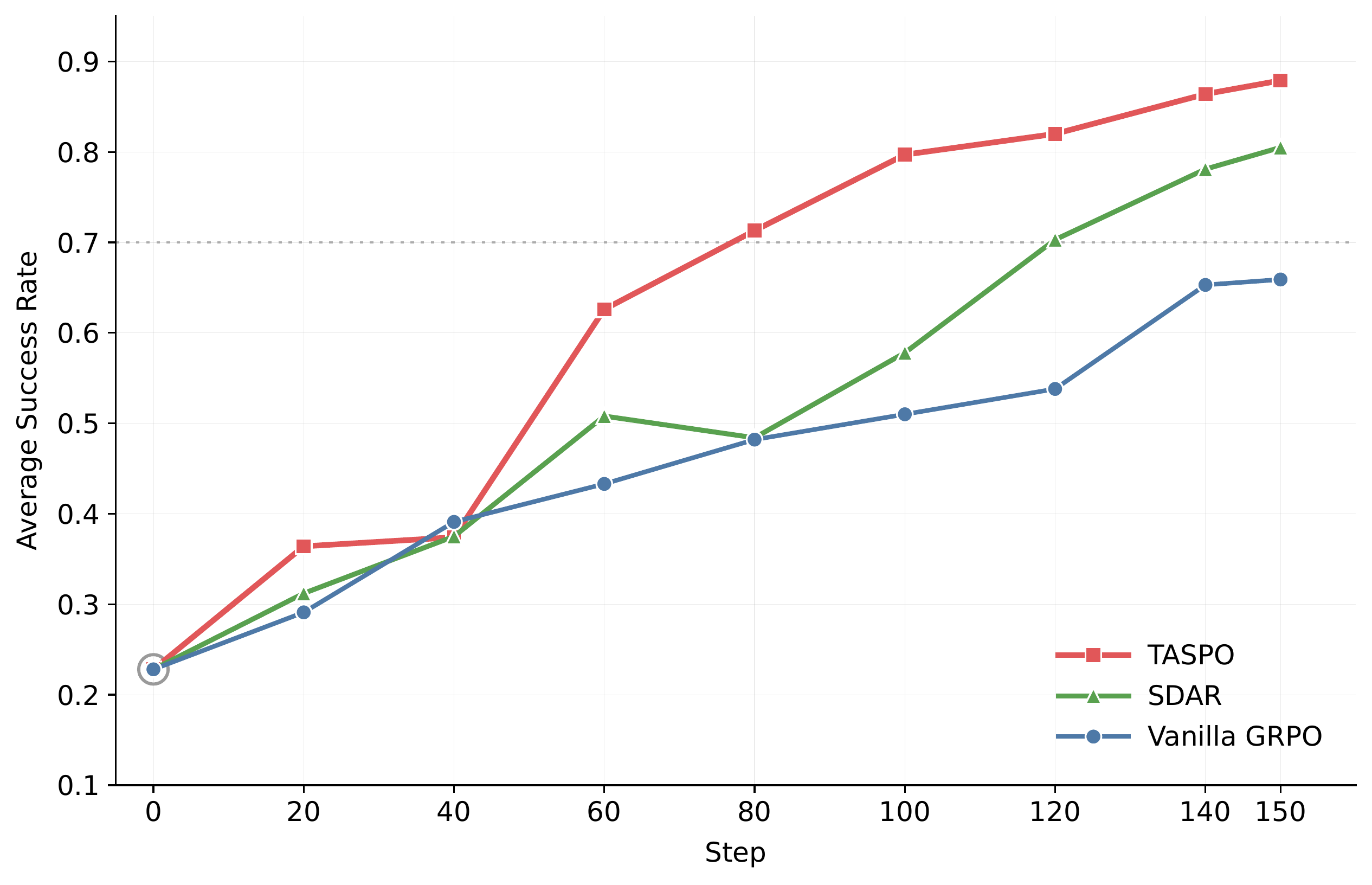}
    % \vspace{-3.0em}
    \caption{Training progress on ALFWorld. TASPO improves faster and reaches a higher final success rate than GRPO and strong baseline SDAR.}
    \label{fig:training-progress}
    \vspace{-1.0em}
\end{wrapfigure}

\textbf{Baselines.}
We compare against the instruction-tuned backbone without post-training(Vanilla), GRPO, and recent methods using training-time skills or privileged supervision, including SDAR, StepOPSD, OPID \citep{lu2026sdar,zhang2026stepopsd,yang2026opid}.
For controlled comparisons, we rerun methods with the same backbone, environment-interaction budget, number of policy updates, and evaluation tasks. Results taken directly from prior work are reported separately for reference, while our main comparisons are based on matched reruns.

\textbf{Training configuration.}
We evaluate Qwen2.5-3B-Instruct, Qwen2.5-7B-Instruct, and Qwen3-1.7B-Instruct. Unless otherwise stated, main experiments use Qwen2.5-3B-Instruct with eight rollouts per task and 150 policy updates.
Task batch sizes are 16 for ALFWorld and WebShop and 128 for Search-QA, with maximum interaction horizons of 50, 15, and 4, respectively. We use a learning
rate of $10^{-6}$ and report averages over three training seeds. TASPO uses a frozen analyzer deepseek-V4-pro to construct and align privileged guidance, with $\tau=[\cdot]$ and $\epsilon_w=[\cdot]$ fixed [across all benchmarks / according to your actual protocol]. Evaluation uses
fixed task and decoding seeds. Complete prompts, hyperparameters, training details, and resource measurements are provided in Appendix~\ref{app:experimental-details}.

\subsection{Main Results}
\label{sec:main-results}

\textbf{Consistent performance gains.}
Table~\ref{tab:main-results} reports results on ALFWorld, Search-QA and WebShop using three backbone models. TASPO consistently outperforms GRPO, improving ALFWorld average success rates by 12.1\%, 11.1\%, and 27.4\% on
Qwen2.5-3B, Qwen2.5-7B, and Qwen3-1.7B, respectively. It also improves Search-QA exact-match accuracy and both WebShop task score and success rate for all three backbones.
These gains support our central premise: process supervision can strengthen outcome-based policy optimization by refining credit within trajectories while preserving the signs of outcome-derived advantages.

\textbf{Improves faster and more effective.}
Figure~\ref{fig:training-progress} shows that TASPO improves faster after the initial exploration stage and reaches a higher final success rate than GRPO and SDAR under the same policy-update budget.
We attribute this to two reasons:
First, Figure~\ref{fig:gradient_geometry} illustrates how an independent distillation
objective can diverge from outcome-based optimization.
TASPO incorporates process supervision through constrained credit redistribution, refining action-level learning signals while preserving outcome-derived advantage signs.
Second, PI is constructed from verified successful siblings within the current rollout group.
Early in training, limited successful experience restricts the availability of applicable guidance, and TASPO falls back to GRPO when none is available.
As the policy improves, more successful rollouts provide additional evidence from which applicable guidance can be constructed.
This can establish a positive feedback loop: policy improvement expands the evidence available for process supervision, which in turn supports more effective credit allocation and further policy improvement.

\subsection{Analysis of Privileged Information}
\label{sec:pi-analysis}

\textbf{PI construction matters.}
Table~\ref{tab:pi-analysis} compares PI construction strategies under the same training protocol. Successful trajectory reuse provides larger gains than generic skills, suggesting the value of concrete interaction evidence.
TASPO's abstracted, trajectory-aligned PI further achieves 86.3\% success, exceeding Nearest Success by 5.2\%.
Its construction combines conditional guidance extraction with target-trajectory matching, addressing source-specific decisions that may be inapplicable to the target's interaction
path. These results support transforming successful experience into decision-applicable supervision before using it to refine outcome-based credit.

\begin{table}[t]
\centering
\caption{
\textbf{PI construction on ALFWorld.}
Alignment denotes matching source guidance to the target trajectory's recorded interaction context.
}
\label{tab:pi-analysis}

\small
\setlength{\tabcolsep}{5pt}
\renewcommand{\arraystretch}{1.05}

\begin{tabular}{lccc}
\toprule
\textbf{PI Construction}
& \textbf{Source}
& \textbf{Alignment}
& \textbf{Success (\%)}
\\
\midrule
GRPO
& --
& --
& 74.2
\\
Generic Skill
& Task skill
& $\times$
& 76.9
\\
Random Success
& Successful trajectory
& $\times$
& 79.2
\\
Nearest Success
& Similar successful trajectory
& $\times$
& 81.1
\\
\rowcolor{gray!15}
\textbf{TASPO}
& \textbf{Abstracted success experience}
& $\checkmark$
& \textbf{86.3}
\\
\bottomrule
\end{tabular}
\end{table}

\textbf{Analyzer robustness.}
We also vary the training-time analyzer while keeping the PI construction procedure and training settings fixed (Table~\ref{tab:analyzer-robustness}). Across the three tested analyzers, PI coverage ranges from 71.6\% to 72.4\%, and success rates from 85.7\% to 86.3\%. These small differences suggest that TASPO's performance
has limited sensitivity to analyzer choice in this setting.

\begin{table}[t]
\centering
\caption{
\textbf{Analyzer robustness on ALFWorld.}
PI coverage and success rates with different training-time analyzers.
}
\label{tab:analyzer-robustness}

\small
\setlength{\tabcolsep}{8pt}
\renewcommand{\arraystretch}{1.05}

\begin{tabular}{lcc}
\toprule
\textbf{Analyzer}
& \textbf{PI Coverage (\%)}
& \textbf{Success (\%)}
\\
\midrule
DeepSeek-V4-Pro
& 72.4
& \textbf{86.3}
\\
GLM
& 71.6
& 85.7
\\
Qwen3.5-35B-A3B
& 72.1
& 86.0
\\
\bottomrule
\end{tabular}
\end{table}

\subsection{Token-Level versus Action-Level Allocation}
\label{sec:credit-analysis}

\textbf{Action-level allocation improves performance and reduces
variability across seeds.}
We compare the two allocation units using identical PI construction
and trajectory alignment, with the same optimization budget and
training configuration. Across three training seeds, action-level
allocation improves performance on all three benchmarks
(Table~\ref{tab:allocation-unit}). Gains are 6.6 percentage points
on ALFWorld and 6.0 points on WebShop, with a smaller 0.5-point
gain on Search-QA. On ALFWorld, the reported cross-seed variation
decreases from 2.9 to 1.4, and successful trajectories require
fewer environment steps on average: 9.2 compared with 11.4.

\begin{table}[t]
\centering
\caption{
\textbf{Effect of credit allocation granularity.}
Results are averaged over three training seeds.
ALF Steps denotes the mean number of environment steps among
successful ALFWorld trajectories; lower is better.
}
\label{tab:allocation-unit}

\small
\setlength{\tabcolsep}{5pt}
\renewcommand{\arraystretch}{1.05}

\begin{tabular}{lcccc}
\toprule
\textbf{Allocation Unit}
& \textbf{ALF Avg.}
& \textbf{Search Avg.}
& \textbf{WebShop Succ.}
& \textbf{ALF Steps}
\\
\midrule
Token-level
& 79.7 $\pm$ 2.9
& 42.9 $\pm$ 1.7
& 72.1 $\pm$ 3.9
& 11.4
\\
\rowcolor{gray!15}
\textbf{Action-level}
& \textbf{86.3 $\pm$ 1.4}
& \textbf{43.4 $\pm$ 1.2}
& \textbf{78.1 $\pm$ 2.6}
& \textbf{9.2}
\\
\bottomrule
\end{tabular}
\end{table}

\begin{figure}[t]
    \centering

    \begin{minipage}[t]{0.48\columnwidth}
        \centering
        \includegraphics[width=\linewidth]{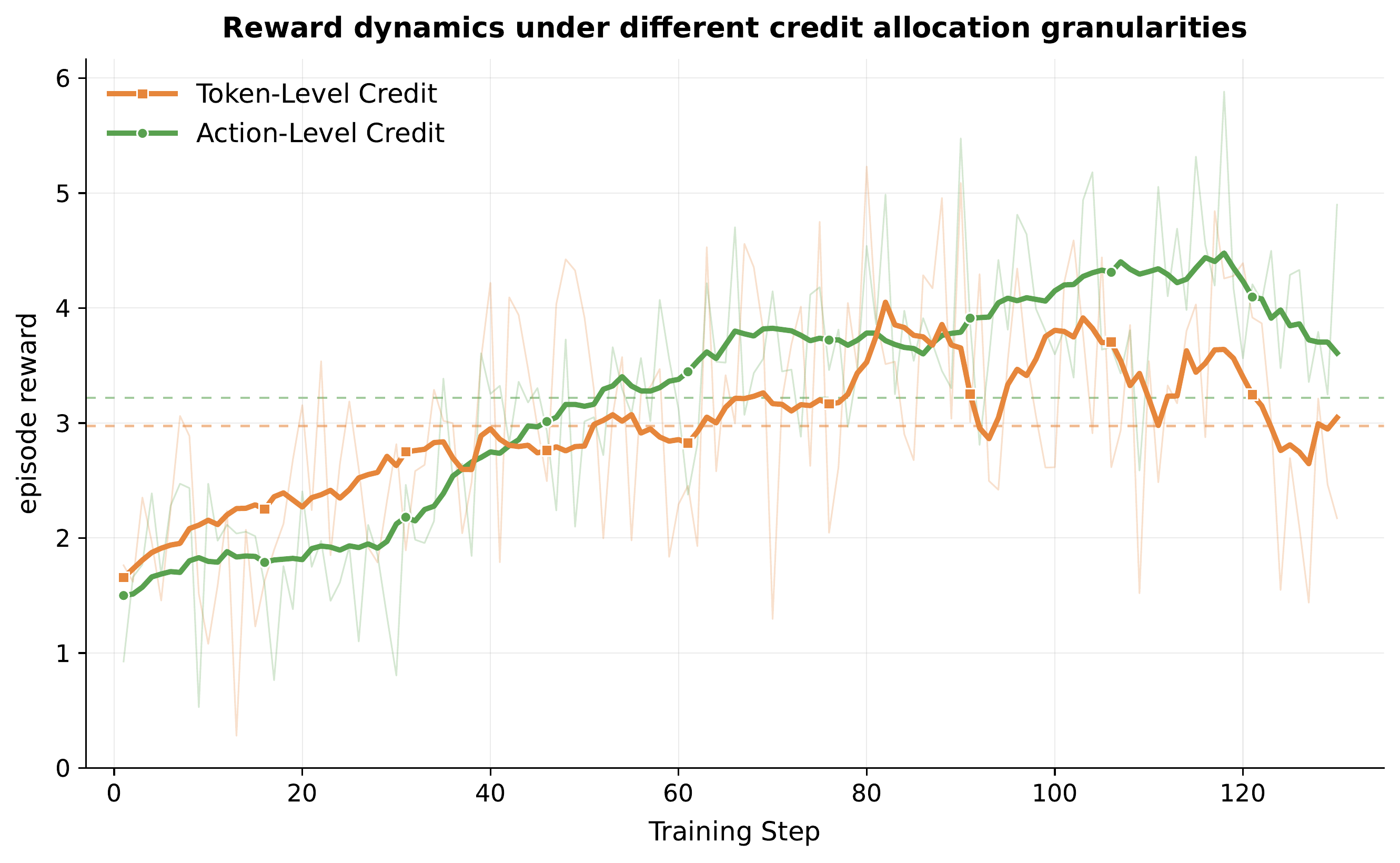}
        \vspace{-0.5em}

        \small\textbf{(a)}
    \end{minipage}
    \hfill
    \begin{minipage}[t]{0.48\columnwidth}
        \centering
        \includegraphics[width=\linewidth]{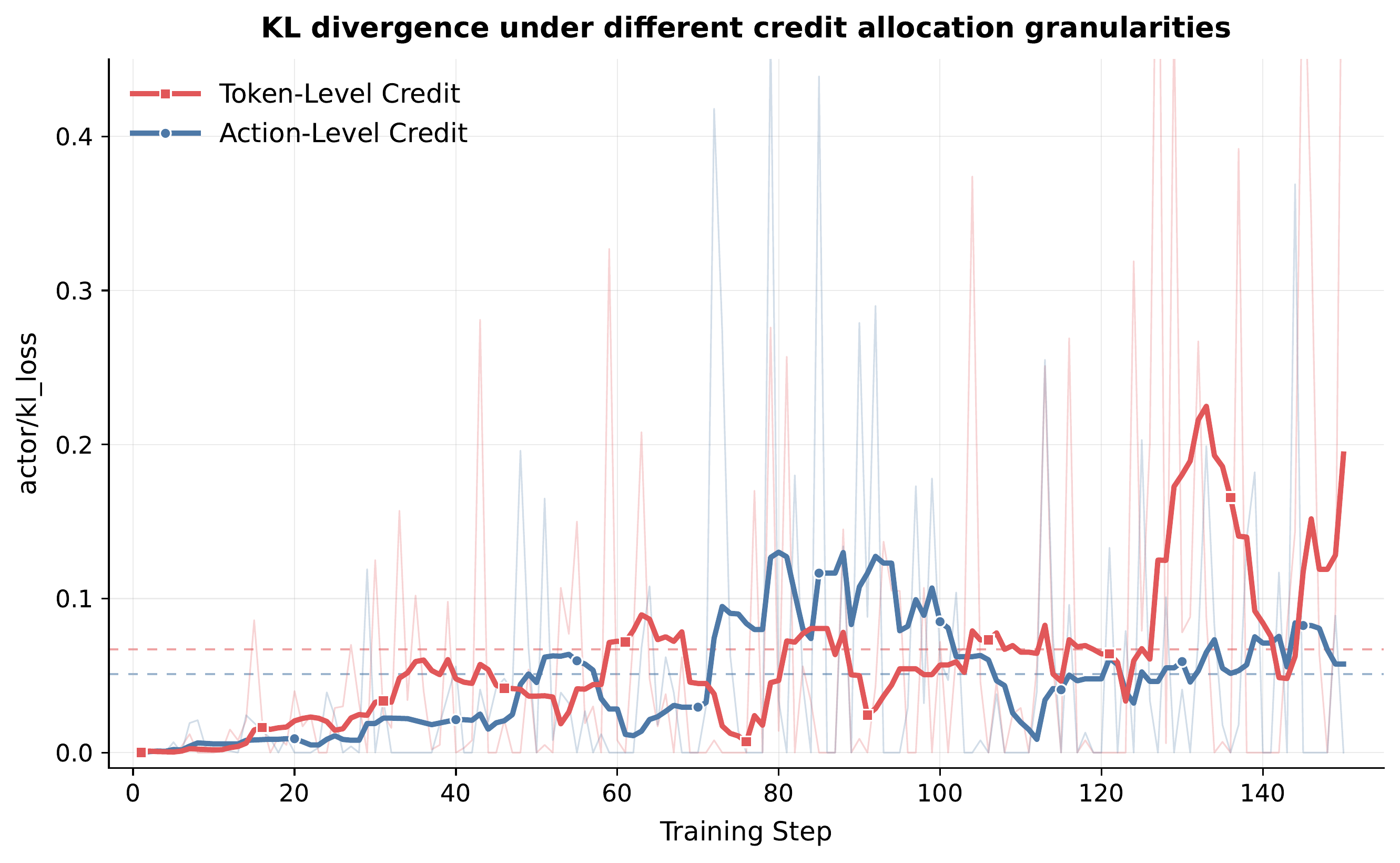}
        \vspace{-0.5em}

        \small\textbf{(b)}
    \end{minipage}

    \caption{
    \textbf{Optimization dynamics under different credit allocation granularities.}
    (a) Action-level allocation achieves smoother reward improvement and a higher final reward compared with token-level allocation.
    (b) Token-level allocation introduces larger policy-update fluctuations, while action-level allocation maintains more stable KL dynamics.
    }
\label{fig:allocation-dynamics}
    \label{fig:optimization-behavior}
\end{figure}

\textbf{Action-level allocation also yields steadier training dynamics.}
Figure~\ref{fig:allocation-dynamics} shows smoother reward progression
and fewer pronounced KL spikes than token-level allocation.
Using a shared action-level credit coefficient prevents individual
tokens within the same executable action from receiving different
PI-derived credit coefficients. The observed dynamics are consistent
with the benefit of reducing these within-action differences
in credit allocation.

\subsection{Credit Reweighting versus Distillation}
\label{sec:reward-analysis}

\begin{wraptable}{r}{0.48\columnwidth}
\centering
\small
\setlength{\tabcolsep}{3pt}
\renewcommand{\arraystretch}{1.05}
\vspace{-2em}
\caption{
\textbf{Supervision strategies on ALFWorld.}
For TASPO, Self + PI denotes the PI-conditioned rollout policy
used for scoring.
}
\label{tab:reward-comparison}

\begin{tabular}{@{}llr@{}}
\toprule
\textbf{Method}
& \textbf{Teacher}
& \textbf{Succ. (\%)}
\\
\midrule
GRPO
& --
& 75.0
\\
OPSD
& Self
& 28.1
\\
OPD
& Qwen2.5-7B
& 76.4
\\
OPD
& Qwen3-32B
& 83.6
\\
GRPO + OPSD
& Self
& 81.7
\\
\rowcolor{gray!15}
\textbf{TASPO}
& \textbf{Self + PI}
& \textbf{87.9}
\\
\bottomrule
\end{tabular}
\end{wraptable}

\textbf{Reweighting outperforms distillation.}
Table~\ref{tab:reward-comparison} compares different ways of
incorporating additional supervision.
Standalone OPSD underperforms GRPO substantially
(28.1\% vs. 75.0\%), whereas GRPO + OPSD improves success to 81.7\%.
TASPO reaches 87.9\%, exceeding the joint objective and the best tested OPD configuration by 6.2\% and 4.3\%, respectively.
These results are consistent with the supervision--credit discrepancy illustrated in Figure~2: an independent distillation objective can favor updates that differ from outcome-based reinforcement. TASPO converts PI into bounded action weights on the original trajectory advantage, preserving its sign while refining credit across decisions. The comparison supports integrating process supervision through outcome-based credit allocation.

\begin{samepage}
\section{Conclusion}
\enlargethispage{2\baselineskip}
This paper we introduced {TASPO} to reconcile process supervision with outcome-based credit in agentic policy optimization. TASPO constructs trajectory-aligned privileged guidance from
successful experience and redistributes action credit while preserving the sign and trajectory mean of the original advantage. Experiments across three benchmarks and three backbones show consistent gains over GRPO. ALFWorld learning curves further show faster improvement. Analyses support its PI construction, action-level allocation, and credit reweighting design.
Process supervision can thus refine local credit while keeping reinforcement grounded in verified outcomes.
\end{samepage}

\clearpage
\bibliography{references}
\bibliographystyle{iclr2027_conference}

\clearpage
\appendix

\counterwithin{table}{section}
\counterwithin{figure}{section}
\counterwithin{equation}{section}

\renewcommand{\thetable}{\Alph{section}.\arabic{table}}
\renewcommand{\thefigure}{\Alph{section}.\arabic{figure}}
\renewcommand{\theequation}{\Alph{section}.\arabic{equation}}

\renewcommand{\theHtable}
{appendix.\Alph{section}.\arabic{table}}
\renewcommand{\theHfigure}
{appendix.\Alph{section}.\arabic{figure}}
\renewcommand{\theHequation}
{appendix.\Alph{section}.\arabic{equation}}

\providecommand{\taspoTODO}[1]
{\textcolor{red}{[TODO: #1]}}
\providecommand{\taspoTBD}
{\textcolor{red}{TBD}}

\section{Method and Implementation Details}
\label{app:method-details}

\subsection{Training Algorithm}
\label{app:training-algorithm}

Algorithm~\ref{alg:taspo} specifies one training iteration.
The analyzer and rollout snapshot are frozen; only the policy is updated.
PI conditions rescoring, while policy optimization and evaluation
use the original contexts.

\begin{algorithm}[H]
\caption{One TASPO training iteration}
\label{alg:taspo}
\small
\begin{algorithmic}[1]
\Require Policy $\pi_\theta$, frozen analyzer, task batch $\mathcal B$,
group size $K$, clipping threshold $c_\Delta>0$,
temperature $\tau>0$, range $0\leq\epsilon_w<1$
\State Freeze the rollout snapshot $\theta_{\mathrm{old}}\gets\theta$
\For{each task $x\in\mathcal B$}
    \State Sample $K$ trajectories with $\pi_{\theta_{\mathrm{old}}}$;
    compute rewards $R_i$ and group-relative advantages $A_i$
    \State Extract $\mathcal E_x$ from verified successful trajectories;
    use $\mathcal E_x=\varnothing$ if none are available
    \For{each target trajectory $\tau_i$}
        \State Initialize $w_{it}\gets1$ for every turn
        \State Exclude evidence supported only by $\tau_i$
        \State Align evidence with the complete target path,
        excluding explicit outcome labels; validate and compose shared $P_i$
        \If{$P_i\neq\varnothing$ and $A_i\neq0$}
            \State Construct paired inputs, parse actions, and rescore
            available turns using $\pi_{\theta_{\mathrm{old}}}$
            \State Collect scorable turns $\mathcal I_i$ using the checks
            in Appendix~\ref{app:method-implementation}
            \If{$|\mathcal I_i|\geq2$}
                \State Compute clipped action means $d_{it}$
                using Eq.~\eqref{eq:app-clipped-action-score}
                \State $\widehat d_{it}\gets d_{it}
                -\operatorname{mean}_{u\in\mathcal I_i}d_{iu}$
                \State $q_{it}\gets
                \tanh(\operatorname{sign}(A_i)\widehat d_{it}/\tau)$
                \State $w_{it}\gets1+\frac{\epsilon_w}{2}
                (q_{it}-\operatorname{mean}_{u\in\mathcal I_i}q_{iu})$
            \EndIf
        \EndIf
        \State Assign detached $\widetilde A_{it}=A_iw_{it}$
        to all policy-generated tokens in each turn
    \EndFor
\EndFor
\State Update $\theta$ on the original policy contexts using
Eq.~\eqref{eq:hierarchical-objective}
\end{algorithmic}
\end{algorithm}

Without a verified success in the group, TASPO retains the original
advantages. It provides no additional process guidance for initial
exploration in this regime.

\subsection{Analyzer Protocol and Prompts}
\label{app:pi-construction}
\label{app:analyzer-prompts}

\textbf{Source evidence and target applicability.}
Extraction identifies source-supported guidance; alignment checks its
applicability to the target path.
Candidates must retain support from another successful trajectory.
Aligned statements cite both source and target evidence and form
one shared $P_i$.

\textbf{Inputs and outcome visibility.}
Both stages receive compact traces as JSON user messages.
Extraction uses verified successes; alignment receives the target task,
all recorded steps, and allowed external evidence.
The target payload excludes \texttt{verified\_success},
\texttt{episode\_reward}, and precomputed advantages.
Actual observations remain unchanged and may reveal task completion;
the exclusion applies to explicit metadata.

\Needspace{8\baselineskip}
\begin{taspoPromptBox}{Prompt 1: Source evidence extraction}

You are a conservative training-time evidence analyst for
a language-model agent.
You receive only environment-verified successful trajectories
for one task. Abstract reusable guidance without inventing
states, observations, actions, or outcomes.

Rules:
\begin{enumerate}[
  leftmargin=1.55em,
  label=\arabic*.,
  itemsep=2pt,
  topsep=3pt,
  parsep=0pt
]
\item Extract task requirements, state-conditioned progress
rules, constraints, and valid alternative paths.
Do not turn one trajectory into a single mandatory plan.

\item Every item must cite exact evidence from at least one
supplied successful trajectory: trajectory id, step,
an environment-accepted action quote, and a pre- or
post-observation quote. Never use an invalid action as
positive source evidence.

\item A statement may be more abstract than its quote,
but it must be entailed by the cited action and
environment observation.

\item Do not assign action scores, advantages, or numeric
weights. Do not diagnose trajectories that are not supplied.

\item If the evidence is insufficient, return fewer items
or an empty list.
\end{enumerate}

Return one JSON object only:
\begin{lstlisting}[style=taspoSchema]
{ "evidence": [{
      "evidence_id": "E1",
      "kind": "requirement|progress|constraint|alternative",
      "statement": "concise natural-language guidance",
      "sources": [{
          "traj_uid": "...", "step": 0,
          "pre_quote": "exact short quote or empty string",
          "action_quote": "exact action quote",
          "post_quote": "exact short quote or empty string"
        }]
}] }
\end{lstlisting}
\end{taspoPromptBox}

\Needspace{8\baselineskip}
\begin{taspoPromptBox}{Prompt 2: Trajectory-level alignment}

Align source-supported guidance to each target's recorded task,
actions, and observations. Explicit terminal reward and success
labels are not supplied. Recorded observations are immutable.

Rules:
\begin{enumerate}[
  leftmargin=1.55em,
  label=\arabic*.,
  itemsep=2pt,
  topsep=3pt,
  parsep=0pt
]
\item For each target, use only its allowed\_evidence items
and ids. These candidates have supporting evidence from
at least one successful trajectory other than the target.

\item Inspect all supplied steps of the realized target trajectory.
Check each item's applicability against the task and recorded
states and actions. Source-supported guidance may still be
inapplicable to this target path.

\item Base applicability on specific interaction evidence.
Do not judge an action solely from an apparent overall outcome
or output inferred terminal labels.

\item Every retained statement must cite allowed source
evidence ids and exact target support from the task,
action, pre-observation, or post-observation.

\item You may reformulate guidance as a target-specific warning
or requirement when supported jointly by the cited source and
target evidence. Never fabricate observations or counterfactual
rollouts. Produce one guidance set for the target trajectory.

\item Do not output action scores, advantages, importance
weights, or numeric updates. If the evidence does not support
alignment, return empty guidance and an abstention reason.
\end{enumerate}

Return one JSON object only:
\begin{lstlisting}[style=taspoSchema]
{ "targets": [{
      "traj_uid": "...",
      "guidance": [{
          "statement": "concise guidance for the shared PI",
          "evidence_ids": ["E1"],
          "target_support": [{
              "step": 2, "field": "task|pre|action|post",
              "quote": "exact short quote"
            }]
        }],
      "abstain_reason": "empty when guidance is retained"
}] }
\end{lstlisting}
\end{taspoPromptBox}

\textbf{Validation.}
Responses must be JSON objects.
Source items require unique nonempty IDs, nonempty statements of at most
800 characters, and references to valid actions in verified successes.
Each reference must match the action quote and a pre- or post-observation
quote.
Alignment requires a known target, a nonempty statement of at most
800 characters, allowed evidence IDs, and a supported target quote.
We check quotes by whitespace-normalized substring matching against
original records. These checks establish traceability; entailment
and applicability remain analyzer judgments.
Rejected items are discarded; empty guidance yields no PI.

\textbf{Context compaction.}
All recorded steps are retained.
After whitespace normalization, overlong fields keep their beginning
and end, separated by \texttt{ ... }.
The character budget $C$ is distributed across the task and steps:
task/action caps are 1,200/320 characters and the minimum observation
budget is 40 characters.
Field minima and JSON overhead make $C$ an approximate request budget.

% Exact per-field budget calculation:
% S = max(1, number_of_steps)
% task = min(1200, max(300, C // 10))
% step = max(180, max(1000, C - task) // S)
% action = min(320, max(80, step // 3))
% observation = max(40, (step - action) // 2)
% At a limit b >= 16, retain (b-5)//2 leading characters
% and b-5-(b-5)//2 trailing characters, separated by " ... ".

\textbf{PI composition and budgeting.}
Accepted statements among the first $G$ candidates are joined as bullets
and shared across turns.
We use token budget $B_{\mathrm{PI}}$ to distinguish it from trajectory
count $B$.
If \texttt{[Privileged Guidance]} and
\texttt{[End Privileged Guidance]} occupy
$B_{\mathrm{wrapper}}<B_{\mathrm{PI}}$ tokens, they enclose the first
$B_{\mathrm{PI}}-B_{\mathrm{wrapper}}$ guidance tokens.
Otherwise, we use the first $B_{\mathrm{PI}}$ guidance tokens alone.
Original contexts and sampled responses are preserved.
Truncation may cut an item: reference checks apply to the original
statement, without guaranteeing completeness of the truncated prefix.

\textbf{Defaults.}
The configuration defaults are $C=12{,}000$, $G=8$,
$B_{\mathrm{PI}}=256$, and an analyzer output limit of 4,096 tokens.
The scoring defaults are $c_\Delta=2.0$, $\tau=0.5$,
and $\epsilon_w=0.4$.

\subsection{Scoring and Loss Implementation}
\label{app:method-implementation}

\textbf{Action spans.}
We locate the first complete \texttt{<action>}, \texttt{<search>},
or \texttt{<answer>} block, matching paired tags case-insensitively
across lines.
After decoding response tokens individually, we include tokens
overlapping the matched span, including both tags.
If no block is found, the same pattern searches the recorded response text.
The complete matched tagged block is tokenized, with or without one
leading space, and located as an exact response-token subsequence.
Failure yields an empty action mask.

\textbf{Paired scoring and valid turns.}
Both passes use the frozen rollout snapshot and original sampled response,
adding $P_i$ only to the privileged context.
A turn enters $\mathcal I_i$ when both inputs are available,
$\mathcal M_{it}$ is complete and nonempty, and every action-token
log probability and difference $\Delta_{itk}$ from
Eq.~\eqref{eq:token-shift} is finite.
Otherwise, we exclude the turn to avoid comparing partial and complete
action scores.

\textbf{Clipped action means.}
For $t\in\mathcal I_i$, we compute
\begin{equation}
d_{it}
=
\frac{1}{|\mathcal M_{it}|}
\sum_{k\in\mathcal M_{it}}
\operatorname{clip}(\Delta_{itk},-c_\Delta,c_\Delta),
\qquad c_\Delta>0.
\label{eq:app-clipped-action-score}
\end{equation}
We apply no small-magnitude filter and average over the complete action.
Both centering operations use the same $\mathcal I_i$:
$d_{it}$ before $\tanh$, and $q_{it}$ when constructing weights.

\textbf{Local and trajectory fallback.}
Unscorable turns keep $w_{it}=1$.
No shared PI, $A_i=0$, or $|\mathcal I_i|<2$ leaves all weights at one.
Otherwise, redistribution proceeds within $\mathcal I_i$ and preserves
the full trajectory mean, as shown in Appendix~\ref{app:theory-credit}.

\textbf{Loss normalization.}
Action-derived weights apply to all generated response tokens, including
reasoning; observations and padding are masked out.
This does not independently assess reasoning correctness.
To implement Eq.~\eqref{eq:hierarchical-objective} with row averaging,
multiply each real turn's token-averaged term by
$N_{\mathrm{row}}/(N_{\mathrm{traj}}T_i)$.
$N_{\mathrm{row}}$ includes padded rows, which contribute zero;
$N_{\mathrm{traj}}$ counts real trajectories.
This implements the nested averaging in the main text.
The balancing factor remains active during fallback and is separate
from the mean-one credit weights.

\section{Derivation and Properties}
\label{app:theory}
\label{app:theory-properties}

\subsection{Weight Derivation and Credit Preservation}
\label{app:theory-credit}

For a fixed trajectory with $A_i\neq0$ and
$n=|\mathcal I_i|\geq2$, let
$q_{it}=\tanh(\operatorname{sign}(A_i)\widehat d_{it}/\tau)$
and $\eta=\epsilon_w/2$.
The weight rule admits the following constrained allocation interpretation:
\begin{equation}
\begin{aligned}
\max_{\{w_{it}\}_{t\in\mathcal I_i}}\quad&
\eta\sum_{t\in\mathcal I_i}q_{it}(w_{it}-1)
-\frac12\sum_{t\in\mathcal I_i}(w_{it}-1)^2\\
\text{s.t.}\quad&
\sum_{t\in\mathcal I_i}(w_{it}-1)=0,
\qquad |w_{it}-1|\leq\epsilon_w.
\end{aligned}
\label{eq:app-allocation-objective}
\end{equation}
The linear term favors larger weights for larger $q_{it}$,
while the quadratic term penalizes departures from uniform credit.

\textbf{Derivation.}
With only the equality constraint, stationarity gives
$w_{it}-1=\eta q_{it}-\lambda$.
Summing over $\mathcal I_i$ gives
$\lambda=\eta\bar q_i$, where
$\bar q_i=n^{-1}\sum_{t\in\mathcal I_i}q_{it}$.
Therefore,
\begin{equation}
w_{it}^*=1+\eta(q_{it}-\bar q_i).
\label{eq:app-weight-solution}
\end{equation}
Since $q_{it},\bar q_i\in[-1,1]$,
$|w_{it}^*-1|\leq2\eta=\epsilon_w$.
The solution already satisfies the box constraints.
Strict concavity therefore makes it the unique optimizer of
Eq.~\eqref{eq:app-allocation-objective}, recovering
Eq.~\eqref{eq:action-weight} without an additional optimization step.

\textbf{Preserved quantities.}
Use Eq.~\eqref{eq:app-weight-solution} on scorable turns and
unit weights elsewhere.
The centered corrections sum to zero within $\mathcal I_i$,
so the complete trajectory has
$\sum_{t=1}^{T_i}w_{it}=n+(T_i-n)=T_i$.
For $0\leq\epsilon_w<1$,
\begin{equation}
1-\epsilon_w\leq w_{it}\leq1+\epsilon_w,
\qquad
w_{it}>0,
\qquad
\frac1{T_i}\sum_{t=1}^{T_i}w_{it}=1.
\label{eq:app-weight-properties}
\end{equation}
Consequently,
\begin{equation}
\begin{aligned}
\operatorname{sign}(\widetilde A_{it})
&=\operatorname{sign}(A_i),&
\frac1{T_i}\sum_{t=1}^{T_i}\widetilde A_{it}
&=A_i,\\
|\widetilde A_{it}-A_i|
&\leq\epsilon_w|A_i|.&&
\end{aligned}
\label{eq:app-credit-preservation}
\end{equation}
The unit-weight fallback satisfies the same properties.
In particular, an action on a negative-advantage trajectory retains
negative credit even when guidance supports it; redistribution
can attenuate that credit without reversing its sign.
These are scalar advantage guarantees.
The parameter-gradient direction also depends on the individual
token gradients and can change under reweighting.

\subsection{Clipped Scores and Relative Support}
\label{app:theory-aggregation}
\label{app:theory-invariance}

\textbf{Bounded token influence.}
Consider a fixed scorable action with $L$ tokens and let
$f_c(z)=\operatorname{clip}(z,-c,c)$, with $c=c_\Delta$.
Since $|f_c(z)|\leq c$,
\begin{equation}
d=\frac1L\sum_{k=1}^{L}f_c(\Delta_k),
\qquad |d|\leq c.
\label{eq:app-clipped-score-bound}
\end{equation}
If only token $j$ changes from $\Delta_j$ to $\Delta'_j$,
clipping is 1-Lipschitz and has range $[-c,c]$, giving
\begin{equation}
|d'-d|
\leq
\frac1L\min\{|\Delta'_j-\Delta_j|,\,2c\}.
\label{eq:app-token-influence-bound}
\end{equation}
Thus, one token changes the action mean by at most $2c/L$,
assuming a fixed action span and unchanged scoring validity.
The bound makes no assumption about which shifts reflect task value.

\textbf{Shared-shift invariance.}
For fixed $\mathcal I_i$, add a constant $b_i$ to every aggregated
action score: $d'_{it}=d_{it}+b_i$.
Then
$d'_{it}-\bar d'_i=(d_{it}+b_i)-(\bar d_i+b_i)=\widehat d_{it}$,
so $q_{it}$ and $w_{it}$ remain unchanged.
This applies to aggregated scores; shared offsets before token clipping
need not cancel because clipping is nonlinear.
The two centerings establish relative support and mean-one weights,
respectively.

\subsection{Compatibility with the Policy Objective}
\label{app:theory-grpo}

Let $r$ be the current-to-rollout policy probability ratio for a
sampled token under its original context.
Write the clipped surrogate to be maximized as
$s_{\mathrm{clip}}(r,A)=
\min\{rA,\operatorname{clip}(r,1-\epsilon_{\mathrm{PPO}},
1+\epsilon_{\mathrm{PPO}})A\}$.
For $A<0$, define the dual-clipped surrogate
$s_{\mathrm{dc}}(r,A)=\max\{s_{\mathrm{clip}}(r,A),c_{\mathrm{dc}}A\}$
with $c_{\mathrm{dc}}>1$; otherwise it equals
$s_{\mathrm{clip}}(r,A)$.
For a fixed detached $w>0$,
\begin{equation}
s_{\mathrm{clip}}(r,wA)=w\,s_{\mathrm{clip}}(r,A),
\qquad
s_{\mathrm{dc}}(r,wA)=w\,s_{\mathrm{dc}}(r,A).
\label{eq:app-grpo-factorization}
\end{equation}
Positive homogeneity of minimum and maximum gives both identities;
the dual-clipped sign condition is also preserved.
They apply equally to the negative-surrogate policy loss.
At fixed $r$, the clipping branch is unchanged.
KL and entropy retain their separate coefficients.

% ============================================================
\section{Experimental Details}
\label{app:experimental}
\label{app:experimental-details}

\subsection{Evaluation Protocols}
\label{app:benchmarks}

Table~\ref{tab:app-protocol} summarizes the evaluation sets.
Results use the prescribed final checkpoint and three
training seeds; held-out tasks are excluded from training
and checkpoint selection.
% \taspoTODO{Confirm exact dataset versions and task IDs,
% evaluation decoding settings and seed values,
% metric normalization, ALFWorld aggregation,
% the meaning of reported error bars,
% and the denominator of PI coverage.}

\begin{table}[!htbp]
\centering
\small
\caption{Training data and evaluation sets.}
\label{tab:app-protocol}
\setlength{\tabcolsep}{4pt}

\begin{tabularx}{\linewidth}{@{}lXXX@{}}
\toprule
\textbf{Domain}
& \textbf{Training data}
& \textbf{Evaluation}
& \textbf{Metrics} \\
\midrule
ALFWorld
& Training split
& 140 seen; 134 unseen tasks
& Success rate \\
WebShop
& Training split
& 128 fixed tasks
& Task score; success rate \\
Search-QA
& NQ and HotpotQA
& Seven datasets listed in the main text
& Exact match \\
\bottomrule
\end{tabularx}
\end{table}

\subsection{Training and TASPO Configuration}
\label{app:optimization}
\label{app:taspo-config}

We use Qwen2.5-3B-Instruct, Qwen2.5-7B-Instruct,
and Qwen3-1.7B-Instruct; controlled analyses use
Qwen2.5-3B-Instruct unless specified otherwise.
Tables~\ref{tab:app-optimization-hparams}
and~\ref{tab:app-taspo-config} collect the training settings.
% \taspoTODO{Verify configuration entries against final-run
% records. Search-QA batch size follows the latest main text;
% other supplied values are retained from the previous
% appendix pending verification.}

\begin{table}[!htbp]
\centering
\small
\caption{Policy optimization and rollout configuration.
Shared settings span all columns.}
\label{tab:app-optimization-hparams}
\setlength{\tabcolsep}{4pt}

\begin{tabular}{@{}lccc@{}}
\toprule
\textbf{Setting}
& \textbf{ALFWorld}
& \textbf{WebShop}
& \textbf{Search-QA} \\
\midrule
Tasks per update & 16 & 16 & 128 \\
Maximum turns & 50 & 15 & 4 \\
Maximum prompt tokens & 2,048 & 4,096 & 4,096 \\
KL coefficient & 0.01 & 0.01 & 0.001 \\
Warmup ratio & 0 & 0 & 0.1 \\
PPO minibatch size & 256 & 64 & 256 \\
Microbatch size per GPU & 16 & 8 & 16 \\
\midrule
Optimizer
& \multicolumn{3}{c}{AdamW} \\
Learning rate
& \multicolumn{3}{c}{$10^{-6}$} \\
Optimizer $(\beta_1,\beta_2)$
& \multicolumn{3}{c}{$(0.9,0.999)$} \\
Weight decay / gradient clipping
& \multicolumn{3}{c}{0.01 / 1.0} \\
PPO clipping / dual-clip coefficient
& \multicolumn{3}{c}{0.2 / 3.0} \\
Policy epochs per batch
& \multicolumn{3}{c}{1} \\
Entropy coefficient
& \multicolumn{3}{c}{0.001} \\
KL estimator
& \multicolumn{3}{c}{Low-variance KL} \\
Rollouts per task / policy updates
& \multicolumn{3}{c}{8 / 150} \\
Rollout temperature / top-$p$
& \multicolumn{3}{c}{1.0 / 1.0} \\
Maximum response tokens
& \multicolumn{3}{c}{512} \\
Number of training seeds
& \multicolumn{3}{c}{3} \\
\bottomrule
\end{tabular}
\end{table}

\begin{table}[!htbp]
\centering
\small
\caption{TASPO-specific settings.}
\label{tab:app-taspo-config}
\setlength{\tabcolsep}{5pt}

\begin{tabular}{@{}ll@{}}
\toprule
\textbf{Setting} & \textbf{Value} \\
\midrule
Default analyzer & DeepSeek-V4-Pro \\
Analyzer temperature / reasoning mode & 0 / disabled \\
Analyzer output limit / workers & 4,096 tokens / 4 \\
Maximum evidence items & 12 \\
PI token budget & 256 \\
Score temperature $\tau$ & 0.5 \\
Weight range $\epsilon_w$ & 0.4 \\
Token-shift threshold $\kappa$ & 0.05\\
\bottomrule
\end{tabular}
\end{table}

\subsection{Baselines and Result Sources}
\label{app:baseline-sources}

Matched runs use the same task-sampling protocol,
rollout group size, update budget, and evaluation tasks.
Each policy generates its own on-policy trajectories.
Reference results retain their original settings.
Table~\ref{tab:app-baseline-sources} records the provenance
supplied in the previous draft.

\begin{table}[!htbp]
\centering
\small
\caption{Provenance of main-table results,
pending final-run verification.}
\label{tab:app-baseline-sources}
\setlength{\tabcolsep}{5pt}

\begin{tabularx}{\linewidth}{@{}lXl@{}}
\toprule
\textbf{Method}
& \textbf{Source}
& \textbf{Comparison} \\
\midrule
Vanilla
& This work
& Matched evaluation \\
GRPO
& This work
& Matched training \\
OPSD
& This work
& Matched training \\
SDAR
& Rerun of \citet{lu2026sdar}
& Matched training \\
StepOPSD
& \citet{zhang2026stepopsd}
& Reference \\
OPID
& \citet{yang2026opid}
& Reference \\
TASPO
& This work
& Matched training \\
\bottomrule
\end{tabularx}
\end{table}

\FloatBarrier  

\end{document}